\documentclass[11pt]{article}

\usepackage[preprint]{acl}

\usepackage{times}
\usepackage{latexsym}

\usepackage[T1]{fontenc}

\usepackage[utf8]{inputenc}

\usepackage{microtype}

\usepackage{inconsolata}

\usepackage{graphicx}

\usepackage{booktabs}
\usepackage{array}  %
\usepackage{amsmath}
\usepackage{xcolor}
\usepackage{CJKutf8}  %
\usepackage{xspace}

\newcommand{\ja}[1]{\begin{CJK}{UTF8}{min}#1\end{CJK}}
\newcommand{\dsname}{HTR Dataset\xspace}
\newcommand{\dsnamefull}{HardToRead (HTR) Dataset\xspace}
\newcommand{\benchname}{SpeechFreq-Yomi\xspace}

\title{Quantifying and Bridging the Spoken--Written Vocabulary Gap \\ in Japanese TTS with Large-Scale Ruby-Guided Speech Synthesis}

\author{
  \textbf{Hikaru Asano\textsuperscript{1,2}},
  \textbf{Yotaro Kubo\textsuperscript{2}},
  \textbf{So Kuroki\textsuperscript{2}}
\\
  \textsuperscript{1}The University of Tokyo,
  \textsuperscript{2}Sakana AI
}

\begin{document}
\maketitle
\begin{abstract}
Text-to-speech (TTS) models learn pronunciation from transcribed speech but must read written text, whose vocabulary is often rare in speech.
This gap is especially severe in Japanese, where many written kanji words are rarely spoken and hard to read from their characters.
We quantify this gap and bridge it with large-scale synthetic speech.
Comparing 247k hours of podcast transcripts with written text, we find nearly twice as many distinct kanji words in writing at equal size, and 9.4\% of written sentences with kanji contain a word seen fewer than ten times in podcasts.
Our pilot experiment shows that difficult readings require hundreds of exposures per word.
We therefore build the \dsnamefull: 139k hours (59M verified utterances) covering 126k speech-rare, writing-frequent word--reading pairs misread by the base model, synthesized by a ruby-guided teacher given each reading.
On \benchname, our 15,150-sentence pronunciation benchmark stratified by transcript frequency, adding the \dsname to podcast fine-tuning raises plain-text TTS accuracy on targeted sentences from 52.9\% to 92.3\% and matches the ruby-guided teacher on words absent from the transcripts.
\end{abstract}

\section{Introduction}
\label{sec:intro}

\begin{figure}[t]
  \centering
  \includegraphics[width=\columnwidth]{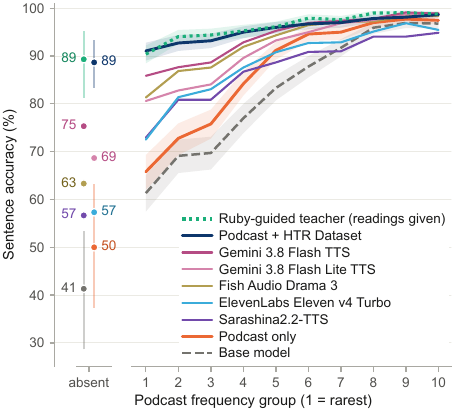}
  \caption{Pronunciation accuracy on \benchname by target-word frequency in podcast transcripts (groups 1--10 from rarest to most frequent, and \emph{absent} words). Podcast-only fine-tuning helps little on rare words. Adding the \dsname raises absent-word accuracy from 50\% to 89\%, matching the ruby-guided teacher (given the readings) and exceeding all other TTS systems on rare words while remaining comparable on frequent words. (\S\ref{sec:main-results}). Bands and bars show 95\% confidence intervals.}
  \label{fig:headline}
\end{figure}

\begin{figure*}[t]
  \centering
  \includegraphics[width=\textwidth]{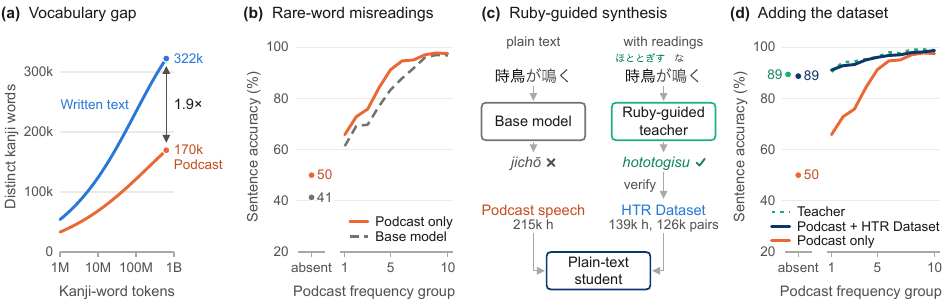}
  \caption[Overview]{Overview. (a)~Distinct kanji words by sample size; at equal size, written text has 1.9$\times$ as many (\S\ref{sec:gap}). (b)~\benchname accuracy by podcast frequency (1~=~rarest; \emph{absent}: unseen); podcast-only fine-tuning helps little (\S\ref{sec:bench}, \S\ref{sec:main-results}). (c)~Given readings, a ruby-guided teacher pronounces written words that are rare in podcasts and misread by the base model; its verified speech and podcast speech train a plain-text student (\S\ref{sec:construction}). (d)~Adding the \dsname moves accuracy toward the teacher (\S\ref{sec:main-results}). Figure~\ref{fig:headline} adds other systems and 95\% intervals; Table~\ref{tab:groups-full} lists the values for all models.}
  \label{fig:teaser}
\end{figure*}

Written language is not simply speech transcribed into text~\citep{Halliday1989-nx}.
Speakers choose words in real time from a compact stock of frequent ones, whereas writers can pause and reach for the exact term~\citep{Chafe1987-kn}.
The result is a large vocabulary gap between spoken and written language~\citep{Biber1988-aw}.
Even children's books contain more rare words than adult conversation~\citep{Hayes1988-xm,Hayes1988-rs}.

Text-to-speech (TTS) models must bridge this gap.
Many recent systems learn pronunciation from speech--transcript pairs~\citep{Du2024-hr}.
In large-scale datasets, these pairs often come from automatically transcribed, largely spontaneous speech~\citep{He2024-je}, so words common in writing but rare in speech receive little supervision.
This uneven lexical coverage hinders pronunciation learning, particularly for speech-rare words whose pronunciation is hard to infer from spelling~\citep{Taylor2019-sa,Sun2023-lu}.
Scaling alone does not resolve this~\citep{Du2025-jp}.
\emph{How well do speech datasets cover the words TTS systems need to read?}

This challenge is especially pronounced in Japanese. A kanji's reading depends on the characters it combines with and sometimes on the context, and the same word may take more than one written form.
Prior work provides synthetic speech for readings of individual commonly used (\emph{jōyō}) kanji~\citep{Liu2026-mg} but does not systematically cover kanji combinations or their readings in context.
The extent of the vocabulary gap and the supervision needed for underrepresented readings remain unclear.

To bridge this gap, we present a large-scale systematic analysis of the spoken--written vocabulary gap in Japanese TTS and experiments aimed at closing it. As shown in Figure~\ref{fig:teaser}, we quantify the spoken--written gap, examine its effect on pronunciation accuracy, and improve rare-word pronunciation through targeted synthetic speech.

We first compare the kanji vocabulary of Japanese podcast transcripts~\citep{PodcastIndex2020-oh,Radford2023-dt} and written text~\citep{Liu2026-eh}.
At equal size, written text has nearly twice as many distinct kanji words, and doubling podcast data adds only 12\% more distinct words (Figure~\ref{fig:teaser}a; \S\ref{sec:gap}).
Words seen fewer than ten times in podcasts appear in 9.4\% of written sentences with kanji.
To measure the gap's effect on TTS, we introduce \benchname, a pronunciation benchmark stratified by podcast frequency (\S\ref{sec:bench}).
Our base model's accuracy~\citep{Hu2026-pn} tracks this frequency, falling to 41\% for words absent from the transcripts (Figure~\ref{fig:teaser}b).
More podcast speech helps little, raising this only to 50\% after 215k hours of fine-tuning and leaving difficult-word misreadings unchanged in our pilot experiment (\S\ref{sec:pilot}).
Targeted supervision is therefore needed.

In that pilot, varying synthetic exposures across 100 difficult words suggests that learning each reading takes roughly 500--1,000 exposures (\S\ref{sec:pilot}), so covering speech-rare words requires a large synthetic corpus.
Building it poses a circular problem: the synthesizer must pronounce words the base model misreads.
We resolve this with a \emph{ruby-guided teacher}, a TTS model given readings alongside text, like ruby annotations beside kanji (Figure~\ref{fig:teaser}c).
It synthesizes the \dsnamefull: 139k hours of verified speech for 126k word--reading pairs frequent in writing, rare in podcasts, and misread by the base model (\S\ref{sec:construction}).

Finally, we train plain-text \emph{students} on podcast speech and the \dsname.
Rare-word accuracy reaches the teacher's level (Figures~\ref{fig:headline} and~\ref{fig:teaser}d): for absent words, it rises from 50\% with podcast-only fine-tuning to 89\% (teacher: 89\%).
On targeted benchmark sentences, accuracy rises from 52.9\% to 92.3\% (\S\ref{sec:main-results}).

Our contributions are threefold:
(i)~\textbf{analysis}, quantifying the vocabulary gap, its link to \benchname errors, and the exposure difficult readings need;
(ii)~\textbf{method}, building the \dsname from gap-driven targets, ruby-guided synthesis, and automatic verification; and
(iii)~\textbf{evaluation}, showing strong rare-word gains and analyzing learning dynamics and remaining errors.

\section{Related Work}
\label{sec:related}

\begin{figure*}[t]
  \centering
  \includegraphics[width=\textwidth]{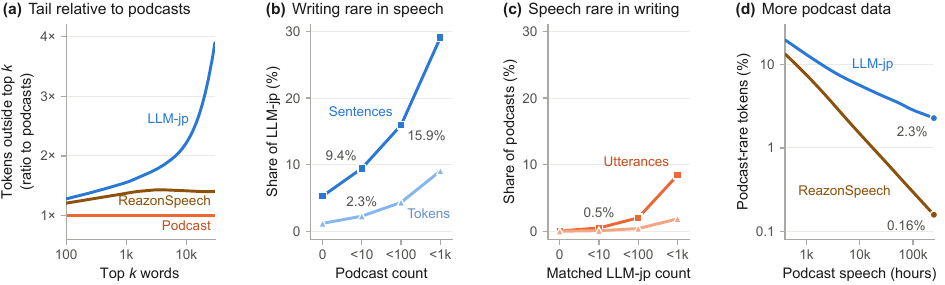}
  \caption[Kanji words in speech and writing]{Kanji words in speech and writing. (a)~Token share outside the $k$ most frequent words, relative to podcasts (90M tokens per corpus). (b)~LLM-jp token and sentence shares below each podcast count threshold (by each sentence's rarest kanji word). (c)~Reverse comparison for podcasts against size-matched LLM-jp. (d)~LLM-jp and TV token shares of words seen fewer than ten times in a random podcast subsample, by subsample size.}
  \label{fig:distribution}
\end{figure*}

\paragraph{Synthetic speech and distillation.}
Synthetic speech augments speech-to-text training data~\citep{Rosenberg2019-jl,Rossenbach2020-md,Fazel2021-sr}, helps correct rare-word errors~\citep{Yamashita2025-bd}, and supports TTS distillation~\citep{Ren2019-ui}.
In Japanese TTS, Sarashina2.2-TTS uses targeted synthetic speech for rare jōyō kanji readings~\citep{Liu2026-mg}, and Ruby-OPD distills ruby guidance into plain-text TTS via on-policy distillation without paired synthetic speech~\citep{Hu2026-ct}.
By contrast, we construct 139k hours of verified synthetic speech targeting words common in writing, rare in podcasts, and misread by the base model.

\paragraph{Transcribed speech corpora and pronunciation.}
Large-scale web corpora~\citep{Li2023-au,He2024-je,Yang2025-cs} supply transcribed speech for modern TTS~\citep{Chen2025-ag,Du2025-jp,Ye2025-df}, yet rare words and ambiguous readings remain difficult~\citep{Taylor2019-sa,Jiang2022-lq,Gao2024-zr}.
Japanese front ends infer readings from dictionaries or annotated text~\citep{Takaoka2018-yk,Sato2022-qb,zhang-2023-pronunciation}, while TTS models can also use phoneme inputs~\citep{Ping2018-eu} or token-level preference optimization~\citep{Kotoge2026-pp}; we instead improve plain-text TTS directly through targeted training data.

\paragraph{Frequency effects and exposure.}
Spoken and written word frequencies differ~\citep{Nohejl2025-xl}, and English TTS errors vary with word rarity~\citep{Borodin2026-fj}; \benchname extends this evaluation to Japanese by grouping word--reading pairs by podcast frequency.
In language models, training frequency predicts factual recall~\citep{Kandpal2022-yk,Mallen2023-on} and lexical or syntactic learning~\citep{Wei2021-hd,Chang2022-tw}, motivating our pilot that varies exposure counts per reading (\S\ref{sec:pilot}).

\section{The Spoken--Written Gap}
\label{sec:gap}

\subsection[Vocabulary Coverage Gap in Spoken and Written Text]{Vocabulary Coverage Gap in Speech and Written Text}
\label{sec:vocabulary-coverage}

\paragraph{Corpora and processing.}
We compare Japanese podcasts from Podcast Index~\citep{PodcastIndex2020-oh} and TV broadcasts from ReazonSpeech~\citep{Yin2023-reazonspeech} with written text from LLM-jp Corpus v4~\citep{Liu2026-eh} (Table~\ref{tab:corpora}).
To preprocess the speech corpora, we extract single-speaker segments from the podcasts and use the supplied segments for TV broadcasts.
We denoise each segment~\citep{Nakata2026-jx} and transcribe it with automatic speech recognition for vocabulary analysis~\citep{Radford2023-dt}, adding ruby annotations to podcast transcripts to mark spoken kanji readings~\citep{Asano2026-ps} (Appendix~\ref{sec:appendix-preprocessing}).
We use the Japanese morphological analyzer Sudachi~\citep{Takaoka2018-yk} to segment each corpus and count occurrences of each dictionary form containing kanji (a \emph{kanji word}).

\paragraph{Written text has a longer tail.}
At 627M kanji-word tokens (the podcast corpus size), an LLM-jp sample contains nearly twice as many distinct words as the podcast corpus (Figure~\ref{fig:teaser}a).
The two speech corpora are much closer: at 90M tokens (the TV corpus size), a podcast sample has 118.9k distinct words and TV 125.5k, against 228.4k for LLM-jp.
Figure~\ref{fig:distribution}a compares the share of tokens outside each corpus's $k$ most frequent words, using 90M tokens per corpus.
As $k$ increases from 1,000 to 30,000, the TV-to-podcast ratio stays near 1.4, while the written-to-podcast ratio rises from 1.6 to 3.9.
The relative gap thus widens toward the rare tail, consistent with the large Jensen--Shannon distances in Table~\ref{tab:corpora}.

\begin{table}[t]
  \centering
  \small
  \setlength{\tabcolsep}{2.5pt}
  \caption{In-dictionary kanji words. JSD is the Jensen--Shannon distance to full LLM-jp.}%
  \label{tab:corpora}
  \begin{tabular*}{\columnwidth}{@{\extracolsep{\fill}}llrrrr@{}}
    \toprule
    Corpus & Domain & Hours & Tokens & Types & JSD \\
    \midrule
    Podcast Index & Podcasts & 247k & 627M & 169.6k & 0.434 \\
    ReazonSpeech & TV & 35k & 90M & 125.5k & 0.407 \\
    LLM-jp v4 & Written & -- & 190.6B & 419.6k & -- \\
    \bottomrule
  \end{tabular*}
\end{table}

\paragraph{Coverage is asymmetric.}
Of the 233,309 distinct kanji words found in at least ten LLM-jp documents (excluding personal names), 51.1\% are absent from the Podcast Index transcripts and 68.8\% occur there fewer than ten times.
Figure~\ref{fig:distribution}b shows that words with fewer than ten podcast occurrences account for only 2.3\% of LLM-jp kanji-word tokens but appear in 9.4\% of LLM-jp sentences containing kanji words.
Raising the cutoff to 100 brings the sentence share to 15.9\%, nearly one in six.
Figure~\ref{fig:distribution}c reverses the comparison. Only 0.5\% of podcast utterances contain a kanji word seen fewer than ten times in a size-matched sample of LLM-jp.
Written text thus routinely draws on words that speech rarely provides, whereas almost every podcast utterance uses only words well attested in writing, even at equal corpus size.

\paragraph{More speech closes the gap slowly.}
Figure~\ref{fig:distribution}d subsamples the podcast corpus and tracks the share of tokens whose word occurs fewer than ten times in the subsample.
For TV speech, the last doubling of podcast data removes 38\% of this share, leaving 0.16\% at 247k hours; for written text, it removes 16\%, leaving 2.3\%.
More podcast data thus quickly covers the vocabulary of other speech but only slowly covers that of writing.

\subsection{\benchname: A Frequency-Stratified Pronunciation Benchmark}
\label{sec:bench}

\paragraph{Benchmark design.}
To assess how this vocabulary gap affects TTS pronunciation, we introduce \benchname.
We stratify LLM-jp kanji words by how often they appear in a random half of the Podcast Index transcripts, using 100 frequency bins on the Zipf scale ($\log_{10}$ occurrences per billion words) plus an \emph{absent} bin.
From each bin, we select 50 word--reading pairs (\emph{items}) and test each item in three sentences (15,150 sentences total).
Within each bin, items are taken in a fixed pseudo-random order from an earlier stratified sample with quotas for proper nouns and minority readings, preferring items whose sentences had already passed review and come from redistributable sources.
For group-level reporting (Figure~\ref{fig:headline} and Table~\ref{tab:main}), the 100 observed bins form ten consecutive groups of ten bins (1,500 sentences; group~1 = rarest, group~10 = most frequent).
Appendix~\ref{sec:appendix-bench} details sampling and reading verification.

\paragraph{Scoring.}
For each model, we synthesize each sentence with a fixed reference voice and transcribe it with Kana-Whisper~\citep{Liu2026-mg}.
Kana-Whisper outputs kana, so models are not penalized when ASR cannot spell a rare kanji word.
Per-bin accuracy is the percentage of sentences whose transcribed target matches an accepted reading; transcription errors can lower measured accuracy.

\paragraph{Pronunciation accuracy by frequency.}
Across the 100 observed bins, accuracy correlates strongly with podcast frequency for all six off-the-shelf systems in Figure~\ref{fig:headline}.
Between the nine highest- and nine lowest-frequency observed bins (right to left in Figure~\ref{fig:headline}), mean accuracy drops by 35 percentage points for the base model (Qwen3-TTS-12Hz-1.7B-Base) and by 13--23 points for the other five systems.
Even Gemini 3.8 Flash TTS, the strongest frontier system we evaluate, scores only 75.3\% on words absent from the podcast sample.

The orange curve in Figure~\ref{fig:headline} shows the base model fine-tuned on 215k hours of transcribed Japanese podcasts (Appendix~\ref{sec:appendix-preprocessing}).
Fine-tuning raises absent-word accuracy from 41.3\% to 50.0\%, but a 32-point gap remains between the nine highest- and nine lowest-frequency observed bins.
Together, these results suggest that the vocabulary gap limits Japanese TTS accuracy unless training data better cover speech-rare written words.

\section{Pilot: How Many Examples Does a Word Need?}
\label{sec:pilot}

The base model's lower accuracy on rare kanji words after fine-tuning on 215k hours of podcasts (\S\ref{sec:bench}) suggests that natural speech alone provides too few examples of their readings.
If we supplement these examples with synthetic speech, how many exposures per word are needed?

\paragraph{Setup.}
To estimate how much synthetic speech to use per word, we compared runs with 0 to 1,024 synthetic exposures per word ($E$) while holding the total number of fine-tuning updates constant.
We used 100 words from an earlier in-house benchmark (Appendix~\ref{sec:appendix-pilot-words}): 30 hard words mostly with non-compositional whole-word readings (\emph{jukujikun}, e.g., \mbox{\ja{時雨}} \emph{shigure} `autumn shower') and 70 verbs, adverbs, and nouns with rare kanji readings.
An LLM generated 64 distinct sentences per word, which a ruby-guided teacher (\S\ref{sec:teacher}) synthesizes in one reference voice (6,400 utterances, 16 hours).

We fine-tuned the base model~\citep{Hu2026-pn} on about 1.3M utterances, substituting synthetic examples for podcast speech and repeating the 64 generated sentences for $E>64$, then scored each target reading in one held-out sentence. %

\begin{figure}[t]
  \centering
  \includegraphics[width=\columnwidth]{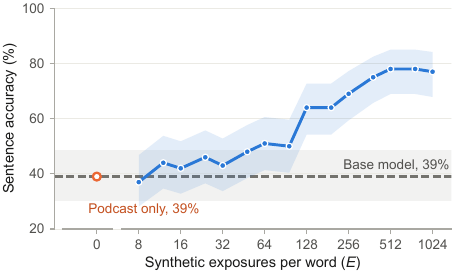}
  \caption[Pilot]{Pronunciation accuracy on 100 words after $E$ synthetic exposures per word. The hollow circle ($E=0$) marks podcast-only fine-tuning; the dashed line shows base-model accuracy. Bands show 95\% Wilson intervals.}
  \label{fig:pilot}
\end{figure}

\paragraph{Results.}
The base model and podcast-only control both achieved 39\% accuracy (Figure~\ref{fig:pilot}).
Accuracy rose to 64\% at 128 exposures and 75\% at 384, then leveled off at 77--78\% from 512 to 1,024.
Relative to the base model, 512 exposures raised accuracy from 33\% to 70\% on hard words and from 41\% to 81\% on rare readings.

\paragraph{Implications.}
The plateau suggests targeting about 512 exposures (and at scale, roughly 512 distinct sentences) per word--reading pair, while prioritizing podcast-rare pairs since frequent ones can be learned from podcast speech (\S\ref{sec:selection}).
The gains from repetition parallel findings on memorization in language models~\citep{Carlini2023-ms,Lu2024-nw,Allen-Zhu2025-tk,Yang2025-ja}, where varied phrasings can also help models learn facts~\citep{Allen-Zhu2024-zq,Lee2026-lh}.

\begin{figure*}[t]
  \centering
  \includegraphics[width=\textwidth]{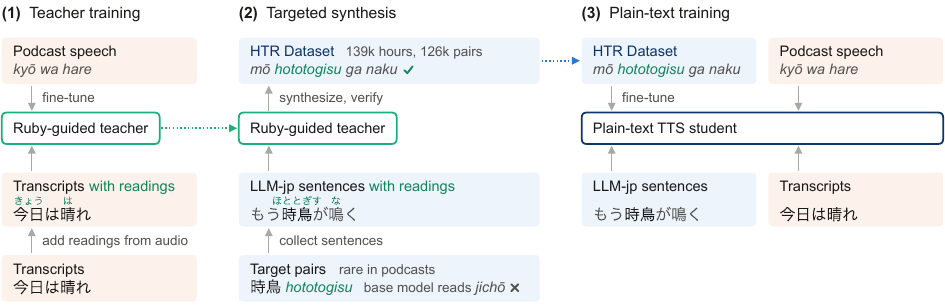}
  \caption{Construction and use of the \dsname. (1)~The teacher trains on podcast speech with audio-inferred readings (\S\ref{sec:teacher}; shown in green with written forms in parentheses). (2)~It synthesizes LLM-jp sentences containing podcast-rare words misread by the base model; verified utterances form the \dsname (\S\ref{sec:selection}--\ref{sec:verification}). (3)~Audio and plain text train the plain-text TTS student (\S\ref{sec:exp-setup}).}
  \label{fig:pipeline}
\end{figure*}

\section{Constructing the \dsname}
\label{sec:construction}

The pilot shows that difficult readings need hundreds of exposures (\S\ref{sec:pilot}).
To supply that supervision for speech-rare words (\S\ref{sec:gap}), we collect written sentences with words the base model misreads, synthesize them with a ruby-guided teacher, and verify the audio to build the \dsname (Figure~\ref{fig:pipeline}).

\subsection{Filtering kanji words for synthesis}
\label{sec:selection}

LLM-jp contains 419,564 distinct kanji words in the Sudachi dictionary (Table~\ref{tab:corpora}).
To keep synthesis practical and efficient, we filter candidates by corpus frequency and then test which readings the base model mispronounces.

\paragraph{Frequency filter.}
Sudachi provides three segmentation granularities: short (mode~A), intermediate (mode~B), and coarsest (mode~C).
For candidate selection, we use mode~B to split long compounds into shorter constituent words and pair each kanji word's written form with its reading.
We select pairs occurring at least 256 times in LLM-jp and at most 1,024 times in podcast transcripts, omitting entries on predefined exclusion lists.
Because pilot accuracy leveled off at 512--1,024 exposures per word (\S\ref{sec:pilot}), the podcast cutoff focuses synthesis on pairs likely to be misread (\S\ref{sec:bench}).
We rank forms by mean relative frequency across the 13 LLM-jp source categories, weighting each category equally to limit the influence of large sources and prioritize broadly used forms during screening and synthesis (Appendix~\ref{sec:appendix-corpus}).

\paragraph{Misreading check.}
Minor spelling differences (such as kana versus kanji spellings in ASR transcripts) can make a word appear rare in the Podcast Index transcripts, and the base model may still read a rare word correctly by combining the usual readings of its constituent kanji.
For each candidate pair, we therefore select five held-out sentences from LLM-jp and synthesize them with the base model from plain text.
We compare the target word's ASR-inferred kana with its reference reading (Appendix~\ref{sec:appendix-filter}) and retain pairs with at least one detected misreading, yielding 106,395 pairs for initial synthesis.

\subsection{Sentence collection}
\label{sec:sentence-collection}

For each selected pair, we collect cleaned and normalized LLM-jp sentences of at most 128 characters where Sudachi assigns the target reading.
Following the pilot, we aim for 600 sentences per pair and deduplicate normalized text across pairs so each sentence is synthesized once.
This initial pass yields 44,291,176 sentences for 106,371 pairs (24 had no usable sentences).
For synthesis, we segment with Sudachi's coarsest mode~C so multi-kanji compounds stay under one ruby tag, keeping annotations where Sudachi and Open JTalk agree and having an LLM reviews the rest.

\subsection{Ruby-guided teacher}
\label{sec:teacher}

Synthesizing the collected sentences with the plain-text base model would reproduce the misreadings detected in \S\ref{sec:selection}.
We therefore train a ruby-guided teacher that takes intended kanji readings alongside the text as privileged information during distillation~\citep{Penaloza2026-if}.

Each training example pairs a transcript, ruby readings, and audio: for each podcast segment (\S\ref{sec:vocabulary-coverage}), we annotate its transcript with readings matching the recorded speech~\citep{Asano2026-ps} (Appendix~\ref{sec:appendix-preprocessing}).
We then fine-tune the base model for one epoch to predict audio codec tokens from the ruby-annotated transcript, so supplied readings guide pronunciation while kanji still inform prosody.
We split sentences into non-overlapping mode~C segments and wrap any segment containing non-kana characters in ruby tags, giving the hiragana reading followed by the written form:
\begin{center}
  \small
  \ja{ただし、}\texttt{<r>}\ja{しちけんしゃ｜質権者}\texttt{</r>}\ja{の}\\
  \texttt{<r>}\ja{しょうだく｜承諾}\texttt{</r>}\ja{を…}
\end{center}
The tags reuse existing special tokens (Appendix~\ref{sec:appendix-teacher}).
To preserve voice cloning, we condition the teacher on a speaker x-vector~\citep{8461375} extracted from another utterance by the same speaker~\citep{Hu2026-pn}.

With explicit readings, the teacher reaches 90.4\% accuracy in the rarest observed group and 89.3\% on items absent from the podcast transcripts, versus 85.9\% and 75.3\% for the best plain-text system, Gemini 3.8 Flash TTS (Figure~\ref{fig:headline}, \S\ref{sec:main-results}).

\subsection{Large-scale synthesis}
\label{sec:synthesis}

Using vLLM-Omni~\citep{vLLM-Omni-Team2026-sc,Yin2026-xb}, we synthesize each sentence once; the initial pass generates 44.3M utterances (about 105k hours).
We subsequently expand synthesis with additional sentences per pair, extra rare and kanji-test targets, and re-synthesized sentences whose reading labels were corrected, bringing the total pool to 81.2M generated utterances.
Before large-scale synthesis, we select 11 clear reference recordings (six male and five female speakers) that yield low character error rates when used as voice references.
Each utterance stores codec tokens (12.5\,Hz, 16 codebooks), 24\,kHz audio, the text and readings, the teacher checkpoint, and the reference recording.

\subsection{Verification and dataset overview}
\label{sec:verification}
\label{sec:dataset}

Filtering unreliable synthetic speech can improve downstream training~\citep{Hu2022-vm}.
We use acoustic rescoring~\citep{Asano2026-ps} to verify that the synthesized audio matches the specified reading, and ASR-based diagnostics to flag synthesis failures such as repetition and truncation.
We also remove superseded originals from label-corrected re-synthesis, utterances outside the 3--30\,s range, and overlaps with evaluation or development sets (Appendix~\ref{sec:appendix-filter}).

After filtering, the \dsname retains 59.3 million utterances (73.0\% of the 81.2M generated), totaling 139,368 hours across 125,539 target word--reading pairs and 11 voices.
In a random sample of 2,000 retained utterances from the initial synthesis release, an independent kana recognizer matched the specified target reading in 90\% of cases.
Compared with podcast speech, synthetic speech scores higher on automatic quality measures and has a narrower range of speaking rates (Appendix~\ref{sec:appendix-dataset}).

\begin{table*}[t]
  \centering
  \scriptsize
  \setlength{\tabcolsep}{2pt}
  \begin{tabular*}{\textwidth}{@{\extracolsep{\fill}}lrrrrrrrrrrr@{}}
    \toprule
    & \multicolumn{7}{c}{\benchname accuracy (\%)} & & \multicolumn{3}{c}{Joyo} \\
    \cmidrule(lr){2-8} \cmidrule(lr){10-12}
    Model & All & Targeted & Non-targ. & Absent & Rarest & Proper & Minority & $\rho$ & Acc. & K-CER$^\dagger_\text{kanji}$\,$\downarrow$ & K-CER$_\text{sent}$\,$\downarrow$ \\
    \midrule
    Qwen3-TTS-1.7B-Base (ja) & 85.1 & 40.7 & 92.4 & 40.7 & 63.9 & 66.0 & 68.5 & 0.94 & 78.3 & 18.52 & 4.32 \\
    Qwen3-TTS-0.6B-Base (ja) & 83.0 & 40.3 & 90.0 & 47.3 & 59.8 & 62.5 & 64.3 & 0.94 & 74.7 & 21.33 & 5.02 \\
    Sarashina2.2-TTS & 87.2 & 72.9 & 89.5 & 56.7 & 72.9 & 74.2 & 82.1 & 0.90 & 90.9 & 7.15 & 4.40 \\
    Gemini 3.8 Flash TTS & 93.9 & 81.8 & 95.8 & 75.3 & 85.9 & 88.0 & \textbf{89.9} & 0.89 & 94.0 & 4.91 & \textbf{2.15} \\
    Gemini 3.8 Flash Lite TTS & 91.5 & 72.3 & 94.7 & 68.7 & 80.6 & 83.2 & 82.4 & 0.93 & 90.0 & 8.09 & 2.61 \\
    Fish Audio Drama 3 & 92.7 & 78.0 & 95.1 & 63.3 & 81.3 & 84.5 & 86.0 & 0.87 & 93.0 & 5.55 & 2.29 \\
    ElevenLabs Eleven v4 Turbo & 88.5 & 68.5 & 91.8 & 57.3 & 72.5 & 74.6 & 80.4 & 0.88 & 91.3 & 7.14 & 3.51 \\
    \midrule
    Qwen3-TTS-1.7B-Base & 82.6 & 39.3 & 89.7 & 41.3 & 61.3 & 60.8 & 67.6 & 0.94 & 77.2 & 19.40 & 5.48 \\
    \quad + Podcast & 86.8 & 52.9 & 92.4 & 50.0 & 65.8 & 72.2 & 67.3 & 0.92 & 82.7 & 14.97 & 4.11 \\
    \quad + Podcast + \dsname & \textbf{95.6} & \textbf{92.3} & \textbf{96.1} & \textbf{88.7} & \textbf{91.1} & \textbf{93.1} & 88.4 & 0.77 & \textbf{95.2} & \textbf{3.60} & 2.24 \\
    \bottomrule
  \end{tabular*}
  \caption{Pronunciation accuracy on \benchname and Joyo Kanji Yomi Benchmark~\citep{Liu2026-mg}. For Qwen3-TTS, (ja) forces Japanese mode; the lower base row uses auto-detection. Mixed-data shows the highest-scoring LR student (matched-LR in Appendix~\ref{sec:appendix-training}). $\rho$: Spearman's correlation across 100 bins. Joyo K-CER ($\times 100$, $\downarrow$): target-span ($\dagger$, capped at 1) and full-sentence kana CER.}
  \label{tab:main}
\end{table*}

\section{Experiments}
\label{sec:experiments}

We fine-tune TTS students on podcast text--audio pairs with and without the \dsname to test whether synthetic examples improve pronunciation of kanji words rarely heard in speech.

\subsection{Setup}
\label{sec:exp-setup}

\paragraph{Training data.}
For natural speech, we filter single-speaker Podcast Index clips (\S\ref{sec:vocabulary-coverage}) by audio quality, Japanese language probability, and Whisper confidence (Appendix~\ref{sec:appendix-preprocessing}), yielding 87.2M utterances (215k hours).

For the mixed-data student, we add 139k hours of verified synthetic speech from the \dsname (\S\ref{sec:verification}), stripping ruby annotations from both corpora so students receive plain text. Together, the corpora total 146.4M utterances (355k hours, 39.3\% synthetic by duration).

\paragraph{Training setup.}
Both students start from \texttt{Qwen3-TTS-12Hz-1.7B-Base}\footnote{\url{https://huggingface.co/Qwen/Qwen3-TTS-12Hz-1.7B-Base}} and predict audio codec tokens.
For both podcast and synthetic speech, we preserve the target voice by conditioning the student on a speaker x-vector from another utterance by the same speaker.
Each update uses 640 utterances (8 H200 GPUs, 16 utterances per GPU, 5 gradient-accumulation steps).
Shuffling synthetic and natural utterances keeps each target's expected exposure rate roughly constant throughout training.
We evaluate final checkpoints and nine evenly spaced intermediate checkpoints (\S\ref{sec:analysis}). Appendix~\ref{sec:appendix-training} details optimization, learning-rate selection, and a matched-LR comparison.

\paragraph{Evaluation.}
On \benchname (\S\ref{sec:bench}), we report pronunciation accuracy for sentences whose benchmark item was targeted for synthesis (\emph{targeted}, 2,136 sentences across 712 items) and for the remaining 13,014 sentences (\emph{non-targeted}).
Non-targeted items can still occur as other words in synthetic sentences (Appendix~\ref{sec:appendix-dataset}).
We also report accuracy on words absent from the podcast transcripts (150 sentences), in the rarest frequency group (group~1; 1,500 sentences), for proper nouns (291), and for minority readings of words with several readings (336).
On the Joyo Kanji Yomi Benchmark~\citep{Liu2026-mg} (13,095 sentences), we report target-reading accuracy and target-span and full-sentence kana CER.

\subsection{Main results}
\label{sec:main-results}

\begin{figure*}[t]
  \centering
  \includegraphics[width=\textwidth]{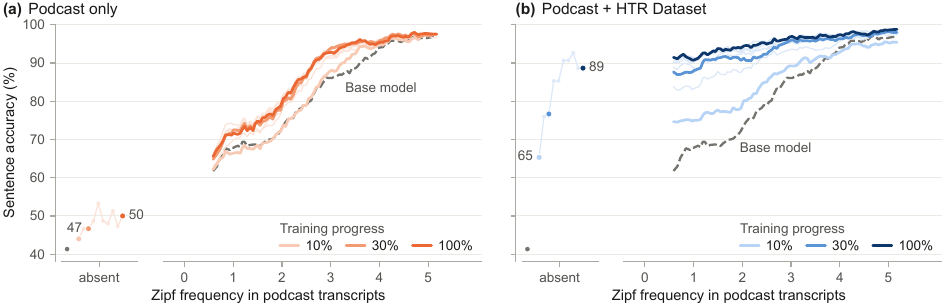}
  \caption{Pronunciation accuracy by podcast frequency at ten checkpoints for (a)~podcast-only training and (b)~training with the \dsname. Curves show nine-bin rolling means; points left of the axis break mark absent words; gray dashed lines show the base model.}
  \label{fig:progress-freq}
\end{figure*}

As Table~\ref{tab:main} and Figure~\ref{fig:headline} show, the student trained on podcasts and the \dsname reaches 95.6\% overall pronunciation accuracy on \benchname, compared with 82.6\% for the base model and 86.8\% for podcast-only training.
The gains over podcast-only training are largest in the long tail that the \dsname targets: accuracy rises by 38.7 percentage points for words absent from podcast transcripts and by 25.3 points in the rarest observed group, compared with 1.2 points in the most frequent group (Appendix Table~\ref{tab:groups-full}).
Accuracy therefore depends less on podcast frequency, with Spearman's $\rho$ falling from 0.92 to 0.77.
Proper nouns and minority readings gain 20.9 and 21.1 points, reaching 93.1\% and 88.4\%.
On non-targeted items, accuracy rises from 92.4\% to 96.1\%, but their words can also occur in synthetic sentences, so this gain does not establish transfer of reading rules to unseen words.
At the podcast-only learning rate ($5\times10^{-5}$), adding an earlier version of the \dsname still raises targeted accuracy from 52.9\% to 78.8\% and absent-word accuracy from 50.0\% to 66.7\% (Appendix~\ref{sec:appendix-training}).

The plain-text student matches the ruby-guided teacher, which receives the intended readings (Figure~\ref{fig:headline}, \S\ref{sec:teacher}): 91.1\% and 90.4\% in the rarest group, and 88.7\% and 89.3\% on absent words.
The checkpoint analysis below shows how the student acquires these readings (\S\ref{sec:analysis}).

\paragraph{Comparison with frontier models.}
On the synthesis-targeted items in \benchname, the student achieves the highest pronunciation accuracy among the plain-text systems in Table~\ref{tab:main} (92.3\%, followed by Gemini 3.8 Flash TTS at 81.8\%), while the two models score similarly on non-targeted items (96.1\% and 95.8\%).
On the Joyo Kanji Yomi Benchmark~\citep{Liu2026-mg}, the student also achieves the highest target-reading accuracy (95.2\%) and the lowest target-span kana CER (3.60\%).
Because the \dsname targets words the base model misreads rather than merely rare ones, its gains extend to readings of commonly used jōyō kanji, exceeding Sarashina2.2-TTS (90.9\%), whose targeted synthesis specifically covers rare jōyō kanji readings.
The student’s whole-sentence kana CER on the same benchmark (2.24\%) lies between those of Gemini 3.8 Flash TTS (2.15\%) and Gemini 3.8 Flash Lite TTS (2.61\%; Table~\ref{tab:main}).
See Appendix~\ref{sec:appendix-results} for more training details and results.

\subsection{Learning rare readings over training}
\label{sec:analysis}

Figure~\ref{fig:progress-freq} shows how rare-word readings develop as the students encounter more training examples.
After the first 30\% of podcast-only training, accuracy on words absent from the stratification sample fluctuates between 46.7\% and 53.3\% over the rest of the epoch.
Because these words occur rarely or never in the podcast training data, more speech of the same kind supplies few examples of them.
With the \dsname, accuracy on these words rises from 65.3\% at 10\% of training to 76.7\% at 30\% and 88.7\% at the final checkpoint, matching the teacher (89.3\%).

Targeted benchmark word--reading pairs occur a median of 697 times in the \dsname (Appendix~\ref{sec:appendix-dataset}).
With the utterances shuffled, a reading with 697 examples receives an expected 70 exposures by 10\% and 210 by 30\%.
Within each frequency group, words with more synthetic examples also show larger gains (Appendix~\ref{sec:appendix-exposure}), suggesting that repeated exposure helps the student memorize individual readings from plain text.

\paragraph{Consistent and remaining errors.}
Items misread in all three sentences drop from 500 for the base model and 341 after podcast-only training to 28 with the \dsname (Appendix~\ref{sec:errors}), showing a sharp reduction in consistent misreadings.
Among remaining misread sentences, 70\% differ from an accepted reading by a single kana (vs.\ 46\% for the base model), often involving voicing (e.g., \ja{横広く} \emph{yokobiroku} as \emph{yokohiroku}) or vowel length, while others reflect Kana-Whisper scorer errors (\S\ref{sec:bench}).

\section{Conclusion}
\label{sec:conclusion}

The spoken--written vocabulary gap limits Japanese TTS pronunciation.
The \dsname, synthesized by a ruby-guided teacher, raises targeted plain-text accuracy from 52.9\% to 92.3\% and matches the teacher on absent words.

\clearpage
\section*{Acknowledgments}

This model was developed with the support from GENIAC (Generative AI Accelerator Challenge), a project implemented by the Ministry of Economy, Trade and Industry (METI) and the New Energy and Industrial Technology Development Organization (NEDO) to strengthen Japan’s generative AI development capabilities and accelerate the adoption of generative AI across society.

\bibliography{custom}

\appendix

\section{Natural Speech Preprocessing}
\label{sec:appendix-preprocessing}

\paragraph{Segmentation and restoration.}
Following existing speech preprocessing practices~\citep{Yaguchi2026-jy}, we split Podcast Index recordings into utterances shorter than 30\,s using voice activity detection.
Speaker diarization~\citep{Bredin2023-oj} identifies utterances with overlapping or multiple speakers, which are removed before speech restoration.
For ReazonSpeech, we use the provided utterance boundaries; these segments can still contain background music, other noise, and speech from additional speakers.
Both corpora undergo speech restoration with SiDON~\citep{Nakata2026-jx} before Whisper transcription~\citep{Radford2023-dt}.

\paragraph{Transcripts and training preparation.}
For podcasts, we run DNSMOS and Whisper on the restored audio and remove segments that fail quality, transcript, or waveform-integrity checks.
For ReazonSpeech, we use the Whisper transcripts produced after restoration for frequency analysis.
The podcast training data additionally carry speaker x-vectors and codec tokens.

\paragraph{Reading annotation.}
Following \citet{Asano2026-ps}, we generate ruby reading labels for the podcast transcripts by rescoring Sudachi, UniDic, and Open JTalk candidates against the audio with a hiragana connectionist temporal classification (CTC) recognizer, using beam search over each segment's candidates.
The selected readings are attached to the transcript segments and serialized in the teacher input format described in Appendix~\ref{sec:appendix-teacher}.

\paragraph{Filtering for student training.}
Preprocessing already restricts podcast segments to a single speaker, DNSMOS OVR $\geq 3.0$, Japanese language probability $\geq 0.85$, and Whisper average log-probability $\geq -0.4$; the frequency analysis uses these segments.
For student training, we further require an average log-probability of at least $-0.2$, a duration of at least 3\,s, and another utterance of at least 3\,s by the same diarized speaker to provide the x-vector reference, leaving 87.2M utterances (215k hours).
Student optimization settings are given in Appendix~\ref{sec:appendix-training}.

\section{Corpus Comparison Details}
\label{sec:appendix-corpus}

\paragraph{LLM-jp source categories.}
We use the Japanese portion of LLM-jp Corpus v4~\citep{Liu2026-eh}, which is released as 14 source subsets.\footnote{\url{https://gitlab.llm-jp.nii.ac.jp/datasets/llm-jp-corpus-v4/-/tree/main/ja}}
We combine SIP HTML with text extracted from SIP PDFs into one category and exclude the OCR-derived part of the SIP PDFs, which LLM-jp-4 pretraining also omits, leaving the 13 categories in Table~\ref{tab:llmjp-categories} for the ranking below.

\begin{table}[t]
  \centering
  \small
  \setlength{\tabcolsep}{4pt}
  \renewcommand{\arraystretch}{1.15}
  \begin{tabular}{@{}l>{\raggedright\arraybackslash}p{0.68\columnwidth}@{}}
    \toprule
    Category & Description \\
    \midrule
    Aozora Bunko & Literary works, mostly out of copyright, from the Aozora Bunko digital library \\
    Common Crawl & Japanese web pages extracted and filtered from all of Common Crawl \\
    e-Gov statutes & Laws and regulations from the e-Gov government portal \\
    FineWeb~2 & Japanese portion of FineWeb~2, a multilingual corpus filtered from Common Crawl \\
    KAKEN & Research project summaries from KAKEN, the database of Grants-in-Aid for Scientific Research \\
    National Diet & Edited session transcripts from the National Diet Library's minutes database \\
    NWC 2010 & Web pages from the HTML archive of the Japanese Web Corpus 2010 \\
    Patents & Patent documents from the Japan Patent Office's published data \\
    SIP web & HTML pages and PDF text from the web corpus built in phase 3 of Japan's Strategic Innovation Promotion Program (SIP) \\
    WARP HTML & HTML pages from mostly public-sector websites listed by the National Diet Library's Web Archiving Project (WARP) \\
    WARP PDF e0 & PDFs from the same websites, created with Word, \TeX, or Ichitaro \\
    WARP PDF e0.2 & PDFs from the same websites, created with PowerPoint, Excel, etc. \\
    Wikipedia & Japanese Wikipedia (20 July 2023 dump) \\
    \bottomrule
  \end{tabular}
  \caption{The 13 LLM-jp Corpus v4 source categories used in this paper. Descriptions are adapted from the corpus documentation.}
  \label{tab:llmjp-categories}
\end{table}

\paragraph{Ranking candidates for the misreading check.}
Within each source category, we count occurrences of each candidate's written form per million mode~B tokens containing kanji (mode~B is the intermediate segmentation granularity of the morphological analyzer Sudachi), including all such tokens in the denominator before filtering candidates.
We average the 13 rates, counting a form's rate as zero in categories where it never occurs.
We check forms in descending order of this mean and, within each form, check readings in descending order of their LLM-jp counts (\S\ref{sec:selection}).

\paragraph{Coverage of written text by speech.}
Table~\ref{tab:coverage-full} extends Figure~\ref{fig:distribution}b to ReazonSpeech and the combined speech corpora, where a word is rare if it is rare in both.
ReazonSpeech is about seven times smaller than the podcast corpus and covers fewer written words; combining them improves coverage only slightly.

\begin{table}[t]
  \centering
  \small
  \setlength{\tabcolsep}{4pt}
  \begin{tabular}{lrrr}
    \toprule
    & Podcast & Reazon & Both \\
    \midrule
    Types absent & 51.1 & 62.3 & 48.7 \\
    Types $<10$ & 68.8 & 79.0 & 68.1 \\
    Tokens $<10$ & 2.3 & 3.5 & 2.2 \\
    Sentences, word absent & 5.3 & 7.3 & 4.5 \\
    Sentences, word $<10$ & 9.4 & 13.8 & 9.0 \\
    Sentences, word $<100$ & 15.9 & 25.7 & 15.7 \\
    \bottomrule
  \end{tabular}
  \caption{Share (\%) of LLM-jp kanji-word types and tokens below a count threshold in each speech corpus, and of sentences whose rarest word falls below it. Types are the 233,309 words in at least ten documents, excluding personal names; tokens and sentences cover all in-dictionary kanji words, including personal names.}
  \label{tab:coverage-full}
\end{table}

\paragraph{Coverage of speech by written text at equal size.}
For the equal-size comparison in Figure~\ref{fig:distribution}c, we thin each LLM-jp kanji-word count with one binomial draw to match the podcasts' 627.0M kanji-word tokens.
Podcast utterances containing a word absent from thinned LLM-jp or seen fewer than 100 or 1,000 times account for 0.06\%, 2.0\%, and 8.4\%, respectively (versus 29.1\% of LLM-jp sentences below 1,000 podcast occurrences).
Below ten and 1,000 occurrences, podcast token shares are 0.10\% and 1.9\% (versus 9.1\% for LLM-jp below 1,000).
Of 70,160 podcast kanji words in at least ten programs (excluding personal names), 0.6\% are absent from thinned LLM-jp and 6.5\% occur fewer than ten times; these type populations come from different-sized corpora and are not strictly comparable.

\section{\benchname Details}
\label{sec:appendix-bench}

\paragraph{Items and stratification.}
There are 244,389 kanji-containing Sudachi mode~B words: the form must be in Sudachi's dictionary, occur in at least ten LLM-jp documents, and not be a personal name; the reading must account for at least 1\% of the form's occurrences and occur at least five times in at least one of five reading sources (Sudachi readings in LLM-jp, podcasts, or ReazonSpeech, and audio-derived readings in podcasts or ReazonSpeech that match a known reading).
We randomly split the Podcast Index programs into two halves and use half~A to assign items to frequency bins.
Of the 100 bins for observed items, one bin covers each count from 1 to 8, the middle bins have equal width, and the high-end bins have equal populations. A separate bin contains items absent from half~A.
Figures and bin-level correlations place each bin at the median half-A Zipf frequency of its items.

\paragraph{Sentences and gold readings.}
From 2.2M candidates, we select three sentences per item from different LLM-jp documents, each with 15--60 characters and no digits.
For regular items, target readings agreed on by Sudachi, UniDic, and Open JTalk are accepted automatically; all other sentences, including those of every proper-noun and minority-reading item, are judged by an LLM (Claude).
Every sentence undergoes at least two independent LLM reviews of its reading and naturalness, with a third adjudicating disagreements.
Sentences from web-crawled and patent sources, whose terms may not permit redistribution, are replaced with sentences from other sources or, when none are available, with LLM-written sentences (465 in total), which pass the same reviews.
For 191 sentences, more than one reading is accepted (e.g., \ja{坂路} \emph{hanro} or \emph{sakamichi}).
The sentences come from Japanese Wikipedia (7,966), Aozora Bunko (4,531), e-Gov statutes (1,957), records of the National Diet (150), and KAKEN research reports (81), plus the 465 LLM-written sentences.

\paragraph{Scoring.}
The Kana-Whisper transcript is aligned to the expected sentence reading, and kana spelling variants are normalized before the target span is compared with the accepted readings.
The scorer also errs: about 3\% of sentences in the most frequent bins are marked wrong, including transcription errors such as \ja{大学} \emph{daigaku} heard as \emph{daikyaku}.

\paragraph{Base-model errors.}
Proper nouns and minority readings remain harder than regular items within the same frequency group.
LLM-written sentences are about as hard as corpus sentences for the same items (67.5\% vs.\ 67.1\% over 155 items).
Some errors combine the usual readings of individual characters (e.g., \ja{村社} \emph{sonsha} read as \emph{murasha}).

\paragraph{Written form.}
Within a frequency group, spellings that omit okurigana (e.g., \ja{取出す} for \ja{取り出す}) score 52 percentage points below standard spellings (Table~\ref{tab:ortho}).
Words usually written in kana are 18 points harder in kanji, whereas spellings with more kana (e.g., \ja{向きあう}) are 20 points easier.
The hard spellings remove reading cues and are rare in ASR transcripts.

\section{Pilot Word List}
\label{sec:appendix-pilot-words}

The pilot (\S\ref{sec:pilot}) targets the following 100 words, each shown with its target reading in hiragana.

\paragraph{Hard words (30).}
\leavevmode\ja{時雨（しぐれ）、雪崩（なだれ）、欠伸（あくび）、陽炎（かげろう）、土筆（つくし）、案山子（かかし）、山車（だし）、海月（くらげ）、心太（ところてん）、蝸牛（かたつむり）、雲雀（ひばり）、啄木鳥（きつつき）、灰汁（あく）、行灯（あんどん）、雪洞（ぼんぼり）、許嫁（いいなずけ）、仲人（なこうど）、玄人（くろうと）、東雲（しののめ）、女将（おかみ）、山葵（わさび）、蒟蒻（こんにゃく）、団扇（うちわ）、暖簾（のれん）、束子（たわし）、杓文字（しゃもじ）、蟷螂（かまきり）、百舌鳥（もず）、竜胆（りんどう）、山茶花（さざんか）}

\paragraph{Rare readings (70).}
Verbs (50): \ja{蔑む（さげすむ）、嘲る（あざける）、唆す（そそのかす）、貪る（むさぼる）、弄ぶ（もてあそぶ）、憚る（はばかる）、慮る（おもんぱかる）、覆す（くつがえす）、遮る（さえぎる）、蘇る（よみがえる）、賄う（まかなう）、培う（つちかう）、償う（つぐなう）、綻ぶ（ほころぶ）、滞る（とどこおる）、嗜む（たしなむ）、窘める（たしなめる）、宥める（なだめる）、拗れる（こじれる）、燻る（くすぶる）、滾る（たぎる）、迸る（ほとばしる）、漲る（みなぎる）、聳える（そびえる）、佇む（たたずむ）、彷徨う（さまよう）、躊躇う（ためらう）、蹲る（うずくまる）、項垂れる（うなだれる）、俯く（うつむく）、足掻く（あがく）、憤る（いきどおる）、悼む（いたむ）、偲ぶ（しのぶ）、労う（ねぎらう）、崇める（あがめる）、貶める（おとしめる）、詰る（なじる）、罵る（ののしる）、嘯く（うそぶく）、囀る（さえずる）、躓く（つまずく）、捗る（はかどる）、穿つ（うがつ）、抉る（えぐる）、抗う（あらがう）、贖う（あがなう）、誂える（あつらえる）、拵える（こしらえる）、設える（しつらえる）}.
Adverbs and adjectival nouns (10): \ja{強か（したたか）、頑な（かたくな）、漸く（ようやく）、悉く（ことごとく）、徐に（おもむろに）、頗る（すこぶる）、遍く（あまねく）、具に（つぶさに）、徒に（いたずらに）、嫋やか（たおやか）}.
Sino-Japanese compounds (10): \ja{建立（こんりゅう）、出納（すいとう）、遊説（ゆうぜい）、会得（えとく）、反故（ほご）、権化（ごんげ）、慟哭（どうこく）、邂逅（かいこう）、逡巡（しゅんじゅん）、畢竟（ひっきょう）}.

\section{Teacher Details}
\label{sec:appendix-teacher}

\paragraph{Input format.}
The ruby tags of \S\ref{sec:teacher} reuse two existing special tokens of the base model.
\begin{center}
  \small
  \begin{tabular}{@{}ll@{}}
    \texttt{<r>} & \texttt{<|object\_ref\_start|>} \\
    \texttt{</r>} & \texttt{<|object\_ref\_end|>}
  \end{tabular}
\end{center}
Segments whose reading is empty or matches their kana spelling remain plain text, and serialization was identical between synthesis and teacher training on all 390,523 sentences checked.

\paragraph{Training.}
The teacher uses 1,000 shards of the podcast data, a learning rate of $5\times10^{-4}$, and a frozen sub-talker module.
An earlier ruby-guided teacher synthesized the training examples for the pilot in \S\ref{sec:pilot}.

\section{Verification and Filtering Details}
\label{sec:appendix-filter}

\paragraph{Recognizers.}
Each utterance is transcribed with Whisper large-v3-turbo and with the hiragana connectionist temporal classification (CTC) recognizer used for reading annotation.
We use the CTC recognizer to rescore candidate readings from Sudachi, UniDic, and Open JTalk against the audio~\citep{Asano2026-ps}.
We do not filter on Whisper's character error rate against the written text: Whisper writes the target in its intended form in only about 16\% of utterances, and 57\% of the utterances with a character error rate above 0.1 nonetheless have correct readings (e.g., the name \ja{平田篤胤} transcribed as \ja{平田厚田根}).

\paragraph{Candidate word scoring.}
For the misreading check in \S\ref{sec:selection}, we align the inferred kana sequence with the reference reading by minimum edit distance to locate the target word in the ASR output.
We compare the normalized kana in the aligned span with the target reading, accepting equivalent kana spellings; differences only in vowel length, including long-vowel marks, are not counted as misreadings.

\paragraph{Verification checks.}
We retain a synthesized utterance if the target word is heard with its specified reading and the whole-sentence reading error rate is at most 0.05 after folding spelling-only differences (e.g., \ja{は}/\ja{わ}).
Repetitions, truncations, and unintelligible outputs are filtered using CTC alignment diagnostics (greedy length ratio $\leq 1.05$ and negative log-likelihood of the reference reading $\leq 2.0$ per character) and Whisper diagnostics (compression ratio $\leq 1.6$, average log-probability $\geq -0.3$, and Japanese language probability $\geq 0.95$); utterances with empty Whisper transcripts are also removed.
We also remove utterances superseded by label-corrected re-synthesis, duplicate sentences within each synthesis file, utterances outside 3--30\,s, and exact matches to evaluation or development sentences.

\section{Dataset Analysis Details}
\label{sec:appendix-dataset}

\paragraph{Statistics.}
The 59,278,856 retained utterances in \dsname have a mean duration of 8.5\,s.
For 70.1\% of utterances, the target reading matches Open JTalk's reading of the written form in isolation; the remaining 29.9\% require other readings (e.g., \ja{下欄} \emph{karan}).

\paragraph{Closing the coverage gap.}
We count occurrences of the 5,050 benchmark items with their selected readings in the filtered TTS training data (\S\ref{sec:bench}).
Including words outside the target spans, \dsname covers 99.8\% of the items at least once.
Among the items in the rarest frequency group, the share that occurs at least ten times in training rises from 38.6\% with podcast data alone to 95.6\% with \dsname, and for items absent from the podcast transcripts from 0\% to 82\%.
Items that were explicitly targeted occur a median of 697 times in \dsname, compared with 58 times in the podcast training data.

\paragraph{Quality.}
Table~\ref{tab:quality} compares automatic quality and reference-voice similarity measures (not used for filtering) on random samples of synthetic and podcast speech.
An independent kana recognizer agrees with the full intended text (character error rate $\leq 0.05$) for 89\% of sampled utterances; some disagreements may reflect recognizer errors on rare words.
Quality is similar across voices.
Speaking rates span a narrower range than in podcasts (2nd--98th percentile 5.1--8.7 morae per second, versus 3.4--10.0).

\begin{table}[t]
  \centering
  \small
  \setlength{\tabcolsep}{4pt}
  \begin{tabular*}{\columnwidth}{@{\extracolsep{\fill}}lrr@{}}
    \toprule
    Measure & Synthetic & Podcast \\
    \midrule
    UTMOS & 4.09 & 3.10 \\
    DNSMOS OVR & 3.47 & 3.34 \\
    Speaker similarity & 0.966 & 0.945 \\
    Median rate (morae/s) & 7.0 & 6.7 \\
    ASR: CER $\leq 0.05$ & 89\% & -- \\
    ASR: target reading heard & 90\% & -- \\
    \bottomrule
  \end{tabular*}
  \caption{Quality of a random sample of utterances from the initial synthesis that passed an earlier, stricter filter ($n=2{,}000$) and of podcast training data ($n=1{,}000$). Speaker similarity is measured against the reference voice. ASR uses an independent kana recognizer; podcast CER is omitted because its references are ASR transcripts.}
  \label{tab:quality}
\end{table}

\section{Student Optimization Details}
\label{sec:appendix-training}

With the sub-talker frozen, we train each student for one epoch (136k updates for podcast-only training and 229k updates for the student trained on podcast speech and \dsname).
Both students use 10\% warmup followed by cosine decay.
Podcast-only training uses a peak learning rate of $5\times10^{-5}$.
For training on podcast speech and \dsname, we use a peak learning rate of $8\times10^{-4}$, as learning new pronunciations required a higher learning rate.

\section{Additional Results}
\label{sec:appendix-results}

Table~\ref{tab:groups-full} gives accuracy by frequency group for the base and fine-tuned models, and Table~\ref{tab:fixed-examples} lists examples of items that training with \dsname fixes.

\begin{table}[!htbp]
  \centering
  \small
  \setlength{\tabcolsep}{3pt}
  \begin{tabular*}{\columnwidth}{@{\extracolsep{\fill}}lrrr@{}}
    \toprule
    Group & Base & Podcast & + \dsname \\
    \midrule
    Absent & 41.3 & 50.0 & 88.7 \\
    1 & 61.3 & 65.8 & 91.1 \\
    2 & 69.1 & 72.8 & 92.7 \\
    3 & 69.7 & 75.8 & 93.2 \\
    4 & 77.0 & 84.2 & 95.0 \\
    5 & 83.3 & 91.2 & 96.0 \\
    6 & 87.8 & 94.6 & 96.7 \\
    7 & 91.7 & 95.0 & 97.0 \\
    8 & 95.9 & 97.0 & 97.9 \\
    9 & 96.9 & 97.7 & 98.1 \\
    10 & 96.8 & 97.5 & 98.7 \\
    \bottomrule
  \end{tabular*}
  \caption{\benchname sentence accuracy (\%) by podcast frequency group (1 = rarest, 10 = most frequent). Podcast: podcast-only training; + \dsname: the mixed-data student in Table~\ref{tab:main}. Each group has 1,500 sentences; the absent group has 150.}
  \label{tab:groups-full}
\end{table}

\subsection{Exposure and written form}
\label{sec:appendix-exposure}

\begin{figure*}[t]
  \centering
  \includegraphics[width=\textwidth]{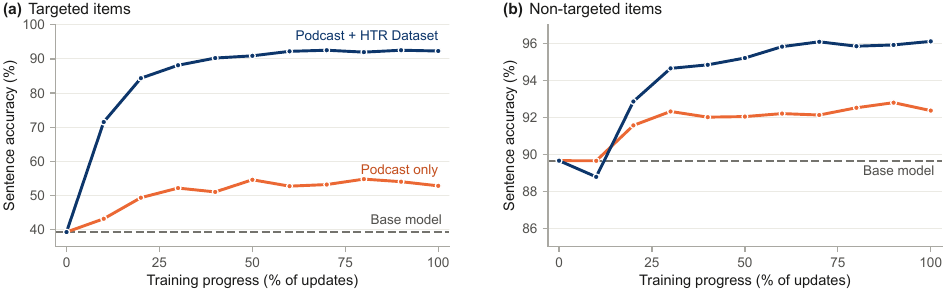}
  \caption{\benchname accuracy at ten evenly spaced checkpoints on (a) targeted and (b) non-targeted items. Shuffling makes training progress proportional to expected target exposure. + \dsname: the student in Table~\ref{tab:main}. Dashed line: base model. The y-axis of (b) is zoomed.}
  \label{fig:learning}
\end{figure*}

\paragraph{Learning over training.}
Targeted accuracy plateaus at 51--55\% after about 30\% of podcast-only training (Figure~\ref{fig:learning}).
With \dsname, it improves until about 70\% (92.6\%), when typical targets have received a few hundred exposures, consistent with the pilot (\S\ref{sec:pilot}).
Non-targeted accuracy initially dips below the base model (88.8\% at 10\% of training, versus 89.7\%) before recovering.

\paragraph{Exposure per item.}
Within frequency groups, podcast-only gains barely depend on training exposure (coefficient of $\log(1+\text{exposure})$: $-0.002$, SE 0.006): items absent from podcast training improve by 7 points, no less than frequent ones.
This suggests a general improvement in reading rather than acquisition of individual words.
With \dsname, synthetic exposure predicts gains (coefficient 0.067--0.072, SE 0.004), consistent with word-specific learning in the pilot.

\paragraph{Written form.}
After podcast-only training, the spelling gaps reported in Appendix~\ref{sec:appendix-bench} narrow from 52 to 50 points for omitted okurigana and remain at 18 points for kana words written in kanji.
With \dsname, these shrink to 4 and 1 points (Table~\ref{tab:ortho}).

\begin{table}[t]
  \centering
  \small
  \setlength{\tabcolsep}{3pt}
  \begin{tabular*}{\columnwidth}{@{\extracolsep{\fill}}lrrr@{}}
    \toprule
    Spelling type (items) & Base & Podcast & + \dsname \\
    \midrule
    Okurigana omitted (36) & $-$52 & $-$50 & $-$4 \\
    Kana word in kanji (89) & $-$18 & $-$18 & $-$1 \\
    More kana (81) & +20 & +15 & +2 \\
    \bottomrule
  \end{tabular*}
  \caption{Accuracy difference from standard spellings within frequency groups, weighted over groups (percentage points). Parentheses give item counts.}
  \label{tab:ortho}
\end{table}

\subsection{Error analysis}
\label{sec:errors}

The three sentences per item distinguish consistent from occasional errors.
Items receiving the same wrong reading in all three sentences fall from 122 for the base model and 84 after podcast-only training to 11 with \dsname.
With \dsname, 381 of the base model's 500 consistently misread items become correct in all sentences: 265 synthesis targets and 116 non-targets that also occur in \dsname (median 324 occurrences, versus 113 in podcast training data).

Five items change from correct in all three sentences to incorrect in all three, compared with 381 changes in the other direction.
One is \ja{不味} \emph{fumi}, read as \emph{mazu}, the reading in the frequent adjective \ja{不味い} \emph{mazui} `bad-tasting', which occurs over a thousand times in \dsname.

Single-sentence outcomes are noisy because synthesis is sampled once, so we favor aggregate comparisons over individual examples.

\begin{table}[!htbp]
  \centering
  \small
  \setlength{\tabcolsep}{4pt}
  \begin{tabular}{llll}
    \toprule
    Item & Reading & Base reading & Target \\
    \midrule
    \ja{銀縁} & \ja{ぎんぶち} & \ja{ぎんえん} & yes \\
    \ja{取手} & \ja{とりで} & \ja{とって} & yes \\
    \ja{赤鼻} & \ja{あかはな} & \ja{あかび} & yes \\
    \ja{木柵} & \ja{もくさく} & \ja{きさく} & yes \\
    \ja{石筆} & \ja{せきひつ} & \ja{いしふで} & yes \\
    \ja{筋張っ} & \ja{すじばっ} & \ja{きんばっ} & yes \\
    \ja{後ろ手} & \ja{うしろで} & \ja{うしろて} & yes \\
    \ja{旗艦} & \ja{きかん} & \ja{たかん} & no \\
    \ja{卒塔婆} & \ja{そとば} & \ja{そつとうば} & no \\
    \ja{甲冑魚} & \ja{かっちゅうぎょ} & \ja{こうぎゅうぎょ} & no \\
    \bottomrule
  \end{tabular}
  \caption{All three sentences are misread by the base and podcast-only models and corrected by the \dsname student ($5\times10^{-5}$, earlier version of \dsname). Base reading: one observed error; non-targets occur elsewhere in synthetic sentences.}
  \label{tab:fixed-examples}
\end{table}

\end{document}